\documentclass[runningheads]{llncs}

\usepackage[T1]{fontenc}
\usepackage{verbatim}
\usepackage{graphicx}
\usepackage{hyperref}
\usepackage{multirow, multicol}
\usepackage{booktabs}
\usepackage{siunitx}
\usepackage{amsmath}
\usepackage{float}
\usepackage{amsfonts}
\usepackage{bm}
\usepackage{marvosym}
\usepackage{ulem}
\usepackage{cite}
\usepackage{array}
\usepackage[table]{xcolor}
\usepackage{arydshln}
\usepackage[flushleft]{threeparttable}

\newcolumntype{L}[1]{>{\raggedright\let\newline\\\arraybackslash\hspace{0pt}}m{#1}}
\newcolumntype{C}[1]{>{\centering\let\newline\\\arraybackslash\hspace{0pt}}m{#1}}
\newcolumntype{R}[1]{>{\raggedleft\let\newline\\\arraybackslash\hspace{0pt}}m{#1}}

\begin{document}

\title{X-Edit: Exact, Explicit, and Explainable Null-Space Editing for Medical Vision Transformers}

\author{
Yuanye Liu\inst{1}\protect\footnotemark[2] \and
Siyuan Zhou\inst{1}\protect\footnotemark[2]  \and
Ke Zhang\inst{2} \and
Lei Li\inst{3} \and
Wei Chen\inst{4} \and
Xiahai Zhuang\inst{1}\protect\footnotemark[5]
}
\authorrunning{Y. Liu et al.}

\institute{
Fudan University, Shanghai, China \and
Johns Hopkins University, Baltimore, USA \and
National University of Singapore, Singapore \and
University of Sydney, Sydney, Australia
}

\makeatletter
\renewcommand\@fnsymbol[1]{%
  \ensuremath{%
    \ifcase#1\or
      *\or
      \dagger\or
      \ddagger\or
      \S\or
      \text{\Letter}\or
      \mathparagraph\or
      \|\or
      **\or
      \dagger\dagger
    \fi
  }%
}
\makeatother
\renewcommand{\thefootnote}{\fnsymbol{footnote}}
\footnotetext[2]{These two authors contributed equally.}
\footnotetext[5]{Corresponding authors: Xiahai Zhuang (zxh@fudan.edu.cn)}

\maketitle

\begin{abstract}
Pre-trained Vision Transformers (ViTs) are increasingly deployed for medical image classification.
However, correcting their inevitable failure cases in dynamic clinical scenarios poses a critical challenge.
Conventional fine-tuning approaches inherently suffer from catastrophic forgetting, severely degrading previously acquired diagnostic capabilities.
Such instability fundamentally compromises clinical safety.
Addressing this vulnerability requires an active, controllable, and reliable intervention mechanism that is both theoretically grounded and inherently interpretable.
To this end, we propose \textbf{X-Edit} (\textbf{eX}act, \textbf{eX}plicit, and \textbf{eX}plainable Editing), an efficient null-space model editing framework.
X-Edit transitions the editing process from iterative gradient-based optimization to a theoretically grounded, closed-form solution.
Specifically,
we first explicitly localize the influential layers via causal tracing governing the erroneous prediction.
Subsequently, we construct an orthogonal null-space projection matrix from a curated anchor set.
By geometrically constraining the exact parameter update strictly within this null space, we provide mathematical guarantees that the intervention rectifies targeted errors without perturbing established diagnostic representations.
Extensive evaluations on six medical imaging benchmarks demonstrate that X-Edit comprehensively suppresses catastrophic forgetting while achieving superior edit success rates.
Our code is available at \url{https://github.com/HenryLau7/X-Edit}.

\end{abstract}

\section{Introduction}

\begin{figure}[t]
    \centering
    \includegraphics[width=\textwidth]{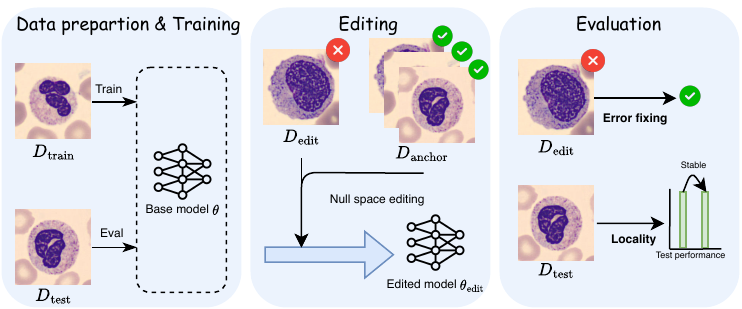}
    \caption{Pipeline of X-Edit. A base model trained on $D_{\text{train}}$ is edited to correct new error samples $D_{\text{edit}}$ using null-space updates constrained by anchor samples $D_{\text{anchor}}$, preserving existing representations. The edited model is then evaluated for error fixing and locality on $D_{\text{edit}}$ and a disjoint test set $D_{\text{test}}$ to ensure stable performance.}
    \label{fig:pipeline}
\end{figure}

Pre-trained Vision Transformers (ViTs)~\cite{A_2021_vit} have demonstrated remarkable efficacy across a wide range of medical image classification tasks~\cite{A_2021_transunet,C_2021MICCAIW_SwinUNetr}.
However, when deployed in dynamic clinical environments, these models inevitably encounter failure cases.
These errors often arise from online data streams containing long-tail pathological variants or domain shifts caused by different imaging equipment~\cite{J_2021_domain_shift,C_2025MICCAI_Bayesmm}.
Continually ensuring the reliability of these models is a critical clinical imperative.
The conventional approach of fine-tuning (FT) the model on these newly encountered error samples typically suffers from catastrophic forgetting, wherein the model loses its previously acquired diagnostic capabilities on correct samples.
In clinical AI, such a phenomenon compromises clinical safety, as a model that forgets common medical knowledge after learning several new cases could not be trusted.
Therefore correcting these errors requires an active, controllable, and reliable intervention that explicitly rectifies targeted failures without collateral damage~\cite{J_2025BMJ_future-ai}.

To address this bottleneck, the concept of \textit{Model Editing} has emerged as a promising post-hoc paradigm~\cite{C_2025ICML_forgetting}.
Recent works have attempted to adapt model editing to Vision Transformers.
LWE~\cite{C_2024NIPS_LWE} proposed a meta-learning approach that trains a hypernetwork to generate sparse binary masks, identifying a subset of parameters to be fine-tuned on failure cases.
Concurrently,
RefineViT~\cite{C_2025NIPS_ViT_edit}
intervened on specific attention heads alongside a newly learned representation projection.
However, these approaches often necessitate computationally intensive optimization pipelines and rely on iterative, gradient-based stochastic refinement.
Crucially, they remain fundamentally misaligned with the strict reliability and interpretability required in clinical AI~\cite{J_2022MIA_inter_survey,J_2023NBE_fairness}, where they lack the theoretical grounding required to provide mathematical guarantees against performance degradation.

Null-space projection is a geometric constraint originally pioneered in continual learning~\cite{J_2019NMI_continual_owm} to strictly isolate parameter updates.
This fundamental concept has been successfully adapted to mitigate catastrophic forgetting in Large Language Models (LLMs)~\cite{C_2023ICLR_MEMIT,C_2025ICLR_alpha_edit},
but its potential for ViTs remains unexplored.

To bridge this gap and fulfill the urgent clinical demand for trustworthy AI,
we propose \textbf{X-Edit} (\textbf{eX}act, \textbf{eX}plicit, and \textbf{eX}plainable Editing), an efficient and scalable null-space model editing framework.
As illustrated in Fig.~\ref{fig:pipeline}, we establish a rigorous post-hoc intervention setting.
Given a base model pre-trained on a standard training set, the editing phase actively rectifies newly encountered error samples ($D_{edit}$) by executing null-space updates constrained by a curated set of anchor samples ($D_{anchor}$).
This design guarantees that the intervention achieves targeted corrections without perturbing previously learned representations.
Finally, the framework includes a comprehensive evaluation phase, assessing the updated model on both the specific edit samples to verify error fixing and a strictly disjoint test set to guarantee diagnostic stability.
Our main contributions are summarized as follows:
\begin{enumerate}
\item We introduce the null-space model editing paradigm into medical image analysis for the first time as far as we know. By reframing the catastrophic forgetting problem as a geometric constraint optimization, we provide a mathematically {controllable} and {safe} method to edit ViT without compromising previously learned diagnostic features.
\item We propose X-Edit, a comprehensive pipeline that seamlessly integrates mechanistic {interpretability} with an exact, closed-form analytical solution for parameter updating. This transparent mathematical design makes our framework highly {scalable} and exceptionally suited for long-term clinical deployment.
\item We comprehensively validate our framework across six diverse medical imaging datasets. Extensive evaluations demonstrate that X-Edit successfully translates its theoretical guarantees into empirical {reliability}, achieving a superior balance between edit success rate and overall test stability compared to existing baselines.
\end{enumerate}

\section{Methodology}\label{sec:method}

\begin{figure}[t]
    \centering
    \includegraphics[width=\textwidth]{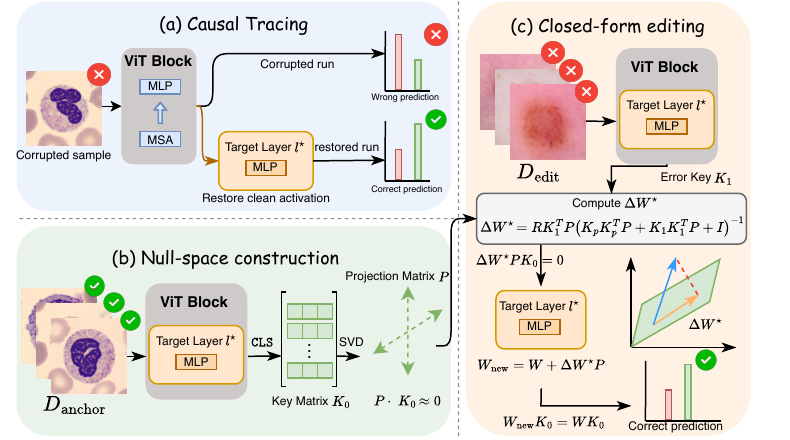}
    \caption{Overview of X-Edit. We first identify influential layers via causal tracing, then construct a null-space constraint from anchor samples, and finally perform a closed-form parameter update to correct model errors while preserving prior knowledge.}
    \label{fig:framework}
\end{figure}

\subsection{Problem Formulation}
Let $f_{\theta}$ denote a pre-trained ViT optimized for a specific medical image classification task, initially trained on a base dataset $D_{train}$. We define an unseen test set $D_{test} = \{(x_i, y_i)\}_{i=1}^N$ purely for evaluation, which strictly does not participate in any training or editing phases.

During deployment, the model may encounter challenging cases, yielding incorrect predictions.
We define these misclassified instances as the edit set $D_{edit} = \{(x_e, y_e)\}_{e=1}^M$, where $f_{\theta}(x_e) \neq y_e$.
The goal of model editing is to find a parameter perturbation $\Delta \theta$ such that the updated model $f_{\theta + \Delta \theta}$ correctly classifies $D_{edit}$. To prevent catastrophic forgetting, we introduce an anchor set $D_{anchor} = \{(x_a, y_a)\}_{a=1}^K$ consisting of accurately predicted samples from $D_{train}$. Formally, our objective is to minimize the error on the edit set while strictly preserving the representations of the anchor set as a proxy for test set stability:
\begin{equation}\label{eq:opt_prob}
\begin{aligned}
    \min_{\Delta \theta} \quad & \mathbb{E}_{(x_e, y_e) \in D_{edit}} [\mathcal{L}(f_{\theta + \Delta \theta}(x_e), y_e)] \\
    \text{subject to} \quad & f_{\theta + \Delta \theta}(x_a) = f_{\theta}(x_a), \quad \forall (x_a, y_a) \in D_{anchor}.
\end{aligned}
\end{equation}
where $\mathcal{L}$ denotes the task-specific loss function.

\subsection{Locating Influential Layers via Causal Tracing}
Inspired by~\cite{C_2022NIPS_causal_tracing}, which demonstrated that feed-forward modules mediate factual associations, we adapt causal tracing to the visual domain to identify the exact location of specific diagnostic knowledge within the ViT.

As shown in Fig.~\ref{fig:framework}(a),
given a medical image $x \in D_{edit}$, we first generate a corrupted version $x_{corrupt}$ by injecting Gaussian noise into its embedding to disrupt the prediction.
We then perform three forward passes:
(1) a \textbf{clean run} with $x$ to cache the clean hidden activations $\{h_l^{clean}\}$ across all transformer blocks;
(2) a \textbf{corrupted run} with $x_{corrupt}$ to obtain the baseline degraded prediction probability;
and (3) a \textbf{restored run}, where the input remains $x_{corrupt}$, but we artificially restore the MLP activation at a specific layer $l$ back to $h_l^{clean}$ during forward propagation.
In the restored run at layer $l$, we compute the Indirect Effect (IE), defined as the magnitude of recovery in the prediction probability of the correct label.
This restoration process is iterated across all transformer blocks.
For each error sample $x \in D_{edit}$, we dynamically identify the top-$k$ layers yielding the highest IE. These identified layers serve as the precise targets for our subsequent null-space parameter editing. For notational simplicity, we collectively denote this sample-specific target set as $l^*$.

\subsection{Null-Space Construction with Anchor Tokens}
According to the linear associative memory interpretation of MLPs~\cite{C_2024ACLf_wilke,C_2025ICLR_alpha_edit}, the weight matrix $W$ of the target layer maps input keys to output values.
As shown in Fig.~\ref{fig:framework}(b), to preserve anchor knowledge, we extract the \texttt{[CLS]} token input activations to the MLP block at layer $l^*$ for all $K$ samples in $D_{anchor}$. Stacking these forms the key matrix $K_0 \in \mathbb{R}^{d_{in} \times K}$, where $d_{in}$ is the input dimension.

Projecting the targeted perturbation $\Delta W$ into the left null space of $K_0$ via a projection matrix $P$ satisfies $\Delta W P \cdot K_0 = 0$.
Consequently, the updated parameters yield:
\begin{equation}\label{eq:null_space}
(W + \Delta W P) K_0 = W K_0 = V_0,
\end{equation}
where $V_0$ represents the original output values.
This strictly guarantees the preserved knowledge is undisrupted.
Computing the null space directly for $K_0$ is computationally prohibitive for large $K$.
Following AlphaEdit~\cite{C_2025ICLR_alpha_edit}, we instead compute it on the non-central covariance matrix $K_0 K_0^\top \in \mathbb{R}^{d_{in} \times d_{in}}$ via SVD:
\begin{equation}\label{eq:SVD}
U \Lambda U^\top = \text{SVD}(K_0 K_0^\top).
\end{equation}
We isolate the null space by removing eigenvectors in $U$ corresponding to non-zero eigenvalues.
To ensure numerical stability, we filter out eigenvectors with eigenvalues greater than a negligible threshold, defining the remaining submatrix as $\hat{U}$.
The orthogonal projection matrix is thus constructed as $P = \hat{U} \hat{U}^\top$. This geometric constraint completely isolates our editing operations from the anchor representations.
Notably, this formulation inherently supports privacy-preserving clinical deployment. Since $P$ relies solely on the pre-computable covariance matrix $K_0 K_0^\top$, the original raw images in $D_{anchor}$ are strictly not required during the online editing phase, circumventing clinical data-availability constraints.

\subsection{Closed-Form Update for Sequential Editing}
To correct the $M$ samples in $D_{edit}$, we extract their input activations at layer $l^*$ to form the key matrix $K_1 \in \mathbb{R}^{d_{in} \times M}$.
Following the paradigm of~\cite{C_2023ICLR_MEMIT,C_2025ICLR_alpha_edit}, we apply several optimization steps to these hidden representations to obtain refined and robust activation targets.
Our goal is to find a perturbation yielding target values $V_1 \in \mathbb{R}^{d_{out} \times M}$ that minimize the classification error.

Under the null-space constraint $\Delta W P$, as in Eq.~\eqref{eq:opt_prob}, we minimize $\|(W + \Delta W P)K_1 - V_1\|^2$, introducing an $L_2$ regularization term $\|\Delta W P\|^2$ to prevent perturbation explosion.
For sequential clinical deployments, we must also ensure new updates do not overwrite previous edits.
Let $K_p$ represent the key matrix of all previously edited samples.
Since past edits are successfully assimilated ($W K_p \approx V_p$), the preservation constraint simplifies to minimizing $\|\Delta W P K_p\|^2$.
The comprehensive objective is:
\begin{equation}
\min_{\Delta W} \left( \|(W + \Delta W P)K_1 - V_1\|^2 + \|\Delta W P K_p\|^2 + \|\Delta W P\|^2 \right).
\end{equation}
Defining the residual error as $R = V_1 - W K_1$ and setting the derivative to zero yields the optimal, closed-form solution:
\begin{equation}\label{eq:solution}
\Delta W^* = R K_1^\top P (K_p K_p^\top P + K_1 K_1^\top P + I)^{-1}.
\end{equation}
The updated weight matrix is $W_{new} = W + \Delta W^*P$.
This analytical solution obviates the need for computationally intensive gradient-based fine-tuning, facilitating post-hoc exceptionally fast interventions.
By structurally embedding $P$ and $K_p$, the framework natively supports scalable sequential editing while mathematically circumventing catastrophic forgetting.

\section{Experiments}

\subsection{Experimental Setup}

\textbf{Baselines.} We evaluate our framework against five methods:
{LWE}~\cite{C_2024NIPS_LWE} (state-of-the-art ViT editing),
{EWC}~\cite{J_2017NAC_EWC} (classical continual learning against forgetting),
{FineTune},
{FineTune+L2 Reg}, and full {Retrain} on the entire dataset (an empirical upper bound for editing efficacy and stability).

\noindent\textbf{Datasets.}
We use five datasets from MedMNIST v2~\cite{medmnistv2} (Blood, Derma, Organa, and Retina, with Path reserved for parameter analysis; resized to $224\times224$) and a liver fibrosis benchmark~\cite{J_2025MIA_merit} (LiFS with two binary tasks: S1--S3 vs. S4 (sub1), and S1 vs. S2--S4 (sub2)).
To prevent data leakage, the edit set exclusively comprises misclassified instances sampled from a disjoint validation split.

\noindent\textbf{Metrics.}
We assess editing performance across four comprehensive metrics:
(1) {Test Performance Drop ($\Delta$ACC)} measures the accuracy degradation on the unseen test set.
(2) {Edit Fix Ratio} represents the proportion of targeted error samples that are successfully corrected.
(3) {Drop Per Edit (DPE)} is a novel metric we propose to quantify the editing trade-off by measuring the collateral test accuracy loss incurred per successfully rectified sample, defined as $\max(0,\Delta\text{ACC} / N_{\text{corrected}})$.
(4) {Edit Time} records the average computational cost required per edit, reflecting the practical efficiency for real-world deployments.

\noindent\textbf{Implementation Details.}
For causal tracing, we analyze each failure case using 32 corrupted runs by injecting Gaussian noise ($\sigma=0.1$) into patch tokens. Influential layers are identified by measuring the \texttt{[CLS]} token recovery score, with the top-3 layers targeted for editing. For null-space construction, we filter eigenvalues below $10^{-2}$ and utilize a default anchor set of 500 samples. The target optimization for $V_1$ consists of 5 iterations with a learning rate of $0.1$. All gradient-based baselines are optimized using the Adam optimizer for 10 epochs to ensure convergence on the edit set. For the LWE baseline, a task-specific hypernetwork is pre-trained for each dataset. All experiments are implemented in PyTorch and executed on NVIDIA GeForce RTX 3090 GPUs.

\begin{table}[t!]
\centering
\caption{Quantitative comparison of editing reliability (measured by test accuracy drop, $\Delta$ACC), effectiveness (Fix Ratio), and efficiency (time per sample) across seven medical image datasets. The symbol $\dagger$ denotes baselines that require access to the original pre-training data.}
\small
\setlength{\tabcolsep}{6pt}
\resizebox{\textwidth}{!}{%
\begin{tabular}{llcccc}
\toprule
\multirow{2}{*}
{\begin{tabular}[c]{@{}l@{}}Dataset\\($N$ = \# Test Cases)\end{tabular}}
& \multirow{2}{*}{Methods} & \multirow{2}{*}{$\Delta$ACC (pp)} & \multirow{2}{*}{Fix Ratio} & DPE & Edit Time \\
& & & & (Drop/Fix) & (Sec/Sample) \\
\midrule

\multirow{6}{*}{\begin{tabular}[c]{@{}l@{}}bloodmnist\\($N$=3,421)\end{tabular}}
& ReTrain$^{\dagger}$ & $\uparrow$ 0.29  & 100.00\% (14/14) & 0.00 & 16.44 \\
\cdashline{2-6}
& FineTune            & $\downarrow$ 0.94 & 100.00\% (14/14) & 0.07 & 1.71 \\
& FineTune+L2         & $\downarrow$ 0.94 & 100.00\% (14/14) & 0.07 & 1.68 \\
& EWC\cite{J_2017NAC_EWC}               & $\downarrow$ 1.02 & 100.00\% (14/14) & 0.07 & 1.85 \\
& LWE\cite{C_2024NIPS_LWE}              & $\downarrow$ 3.27 & 85.71\% (12/14)  & 0.27 & 3.03 \\
\rowcolor{gray!15} &
X-Edit                & $\downarrow$ 0.36 & 100.00\% (14/14) & 0.03 & 0.53 \\
\midrule

\multirow{6}{*}{\begin{tabular}[c]{@{}l@{}}dermamnist\\($N$=2,005)\end{tabular}}
& ReTrain$^{\dagger}$ & $\downarrow$ 0.30 & 100.00\% (93/93) & 0.00 & 1.47 \\
\cdashline{2-6}
& FineTune            & $\downarrow$ 4.04 & 100.00\% (93/93) & 0.04 & 0.35 \\
& FineTune+L2         & $\downarrow$ 4.04 & 100.00\% (93/93) & 0.04 & 0.52 \\
& EWC~\cite{J_2017NAC_EWC}                 & $\downarrow$ 3.84 & 100.00\% (93/93) & 0.04 & 0.56 \\
& LWE~\cite{C_2024NIPS_LWE}                  & $\downarrow$ 28.38& 82.80\% (77/93)  & 0.37 & 2.74 \\
\rowcolor{gray!15} &
X-Edit                & $\downarrow$ 1.00 & 98.92\% (92/93)  & 0.01 & 1.31 \\
\midrule

\multirow{6}{*}{\begin{tabular}[c]{@{}l@{}}organamnist\\($N$=17,778)\end{tabular}}
& ReTrain$^{\dagger}$ & $\uparrow$ 0.03   & 100.00\% (3/3)   & 0.00 & 239.62 \\
\cdashline{2-6}
& FineTune            & $\downarrow$ 0.03 & 100.00\% (3/3)   & 0.01 & 31.83 \\
& FineTune+L2         & $\downarrow$ 0.03 & 100.00\% (3/3)   & 0.01 & 24.57 \\
& EWC~\cite{J_2017NAC_EWC}                 & $\downarrow$ 0.06 & 100.00\% (3/3)   & 0.02 & 27.96 \\
& LWE~\cite{C_2024NIPS_LWE}                  & $\uparrow$ 0.06   & 100.00\% (3/3)   & 0.00 & 54.00 \\
\rowcolor{gray!15} &
X-Edit                & $\uparrow$ 0.23   & 100.00\% (3/3)   & 0.00 & 0.53 \\
\midrule

\multirow{6}{*}{\begin{tabular}[c]{@{}l@{}}retinamnist\\($N$=400)\end{tabular}}
& ReTrain$^{\dagger}$ & $\uparrow$ 1.50   & 100.00\% (38/38) & 0.00 & 0.63 \\
\cdashline{2-6}
& FineTune            & $\downarrow$ 3.50 & 100.00\% (38/38) & 0.09 & 0.24 \\
& FineTune+L2         & $\downarrow$ 3.50 & 100.00\% (38/38) & 0.09 & 0.45 \\
& EWC~\cite{J_2017NAC_EWC}                 & $\downarrow$ 4.00 & 100.00\% (38/38) & 0.11 & 0.51 \\
& LWE~\cite{C_2024NIPS_LWE}                 & $\downarrow$ 9.25 & 71.05\% (27/38)  & 0.34 & 3.62 \\
\rowcolor{gray!15} &
X-Edit                & $\downarrow$ 0.50 & 100.00\% (38/38) & 0.01 & 0.91 \\
\midrule

\multirow{6}{*}{\begin{tabular}[c]{@{}l@{}}LiFS-subtask1\\($N$=141)\end{tabular}}
& ReTrain$^{\dagger}$ & $\downarrow$ 11.35& 90.91\% (10/11)  & 1.14 & 1.34 \\
\cdashline{2-6}
& FineTune            & $\downarrow$ 29.79& 100.00\% (11/11) & 2.71 & 0.45 \\
& FineTune+L2         & $\downarrow$ 29.79& 100.00\% (11/11) & 2.71 & 0.64 \\
& EWC~\cite{J_2017NAC_EWC}                 & $\downarrow$ 31.21& 100.00\% (11/11) & 2.84 & 0.74 \\
& LWE~\cite{C_2024NIPS_LWE}                 & $\downarrow$ 33.33& 54.55\% (6/11)   & 5.56 & 2.29 \\
\rowcolor{gray!15} &
X-Edit                & $\downarrow$ 5.68 & 100.00\% (11/11) & 0.52 & 0.92 \\
\midrule

\multirow{6}{*}{\begin{tabular}[c]{@{}l@{}}LiFS-subtask2\\($N$=141)\end{tabular}}
& ReTrain$^{\dagger}$ & $\downarrow$ 1.42 & 50.00\% (4/8)    & 0.36 & 1.92 \\
\cdashline{2-6}
& FineTune            & $\downarrow$ 56.74& 100.00\% (8/8)   & 7.09 & 0.56 \\
& FineTune+L2         & $\downarrow$ 56.74& 100.00\% (8/8)   & 7.09 & 0.73 \\
& EWC~\cite{J_2017NAC_EWC}                 & $\downarrow$ 56.03& 100.00\% (8/8)   & 7.00 & 0.86 \\
& LWE~\cite{C_2024NIPS_LWE}                 & 0.00              & 12.50\% (1/8)    & 0.00 & 1.19 \\
\rowcolor{gray!15} &
X-Edit                & $\downarrow$ 4.25 & 100.00\% (8/8)   & 0.53 & 0.18 \\

\bottomrule
\end{tabular}
 }
\label{tab:main_results}
\vspace{-1.5em}
\end{table}%

\subsection{Results}

\noindent\textbf{Main results.}
A clinically viable model editing framework must reliably correct newly encountered errors without degrading the established diagnostic capabilities. As demonstrated in Table.~\ref{tab:main_results}, unconstrained gradient-based methods, including FineTune, FineTune+L2, and EWC, suffer from severe catastrophic forgetting.
This degradation is particularly devastating in complex, real-world clinical task liver fibrosis staging.
For instance, on the LiFS-subtask2 dataset, standard FT leads to a severe performance collapse, with test accuracy dropping by 56.74 percentage points.
Similarly, the SOTA ViT editing baseline, LWE, exhibits extreme instability, incurring drops of 28.38 and 33.33 percentage points on DermaMNIST and LiFS-subtask1, respectively.
In contrast, X-Edit demonstrates exceptional reliability.
It incurs near-zero performance drops across all MedMNIST datasets and restricts the degradation on the highly challenging LiFS tasks to merely 5.68 and 4.25 percentage points, respectively, remarkably approaching the empirical upper bound established by ReTrain baseline.

Beyond mere preservation of test stability, the intervention must successfully assimilate the targeted corrections.
While standard FT naturally achieves a 100\% Fix Ratio by aggressively overwriting parameters, it destroys the base knowledge.
Conversely, LWE frequently struggles to learn new corrections on hard cases, yielding poor Fix Ratios (54.55\% on LiFS-subtask1 and 12.50\% on LiFS-subtask2).
X-Edit overcomes this dilemma, consistently achieving near 100\% Fix Ratios across almost all benchmarks.
Furthermore, the DPE metric explicitly quantifies this trade-off by measuring the collateral damage inflicted on the test set per successfully rectified sample.
X-Edit achieves the lowest Drop/Fix ratio among all methods.
Furthermore, leveraging its closed-form solution, X-Edit demonstrates high computational efficiency, consistently requiring significantly less editing time than the SOTA ViT editing baseline, LWE.

\begin{figure}[t]
    \centering
    \includegraphics[width=\textwidth]{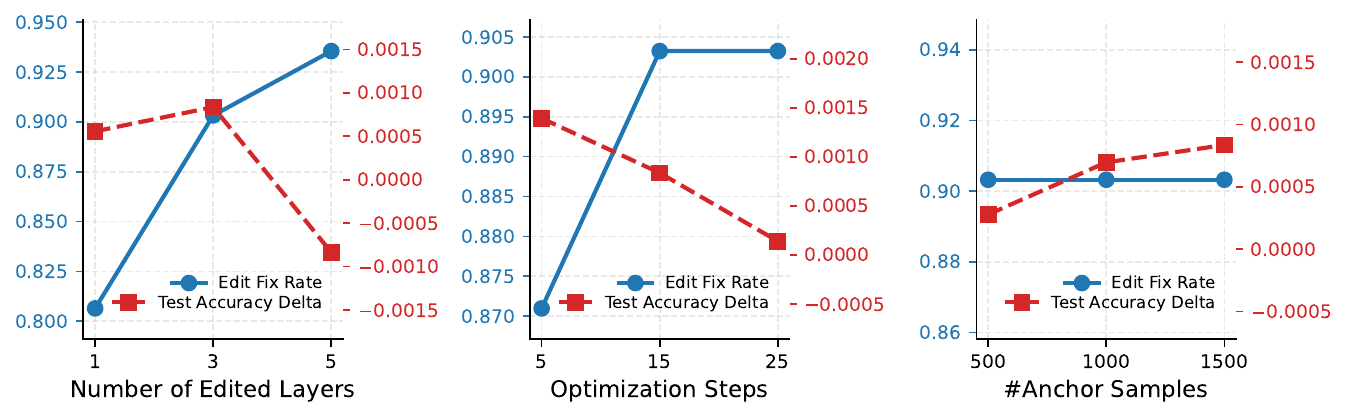}
    \caption{Parameter analysis for the number of edited Layers, optimization steps and size of anchor sets.}
    \label{fig:para_anal}
    \vspace{-1.5em}
\end{figure}

\noindent\textbf{Parameter Analysis.}
We evaluate the sensitivity of X-Edit to three key hyper-parameters on PathMNIST. As illustrated in Fig.~\ref{fig:para_anal}, a trade-off exists between the Edit Fix Ratio and Test Performance Drop:
(1) Increasing the number of edited layers or optimization steps enhances the fix rate but exacerbates test accuracy degradation.
(2) Conversely, expanding the anchor set size reinforces stability by tightening the geometric null-space constraints.
These findings empirically validate our theoretical design, highlighting the balance between acquiring new knowledge and preserving established diagnostic representations.

\section{Conclusion}
In this paper, we proposed \textbf{X-Edit}, a null-space model editing framework designed to rectify errors in medical ViTs without catastrophic forgetting.
By integrating causal tracing for mechanistic interpretability with a mathematically transparent, closed-form analytical solution, X-Edit enables exact and efficient post-hoc interventions. Extensive evaluations demonstrate that X-Edit's geometric constraints successfully translate into empirical reliability.
By delivering \textbf{eXact} updates, \textbf{eXplicit} error correction, and an \textbf{eXplainable} mechanism,
X-Edit provides a step toward controllable post-hoc editing for medical ViTs and can support long-term clinical model adaptation.

\begin{credits}
\subsubsection{\ackname}
This work was funded by the Science and Technology Commission of Shanghai Municipality (25TS1412100), the Noncommunicable Chronic Disease-National Science and Technology Major Project (2026ZD0555800/2026ZD0555802), and the National Natural Science Foundation of China (62372115).

\subsubsection{\discintname}
The authors have no competing interests to declare that are
relevant to the content of this article.
\end{credits}

\bibliographystyle{splncs04}
\bibliography{string,refs}

@inproceedings{C_2022NIPS_causal_tracing,
  title={Locating and editing factual associations in gpt},
  author={Meng, Kevin and Bau, David and Andonian, Alex and Belinkov, Yonatan},
    booktitle = NIPS,
  year={2022}
}

@inproceedings{C_2025ICLR_alpha_edit,
  title={AlphaEdit: Null-Space Constrained Knowledge Editing for Language Models},
  author={Fang, Junfeng and Jiang, Houcheng and Wang, Kun and Ma, Yunshan and Jie, Shi and Wang, Xiang and He, Xiangnan and Chua, Tat-Seng},
    booktitle = ICLR,
  year={2025}
}

@misc{A_2021_vit,
      title={An Image is Worth 16x16 Words: Transformers for Image Recognition at Scale}, 
      author={Alexey Dosovitskiy and Lucas Beyer and Alexander Kolesnikov and Dirk Weissenborn and Xiaohua Zhai and Thomas Unterthiner and Mostafa Dehghani and Matthias Minderer and Georg Heigold and Sylvain Gelly and Jakob Uszkoreit and Neil Houlsby},
      year={2021},
      eprint={2010.11929},
}

@misc{A_2021_transunet,
      title={TransUNet: Transformers Make Strong Encoders for Medical Image Segmentation}, 
      author={Jieneng Chen and Yongyi Lu and Qihang Yu and Xiangde Luo and Ehsan Adeli and Yan Wang and Le Lu and Alan L. Yuille and Yuyin Zhou},
      year={2021},
      eprint={2102.04306},
}

@inproceedings{C_2021MICCAIW_SwinUNetr,
  title={Swin unetr: Swin transformers for semantic segmentation of brain tumors in mri images},
  author={Hatamizadeh, Ali and Nath, Vishwesh and Tang, Yucheng and Yang, Dong and Roth, Holger R and Xu, Daguang},
  booktitle={MICCAI brainlesion workshop},
  pages={272--284},
  year={2021},
}

@inproceedings{C_2025MICCAI_Bayesmm,
  title={BayeSMM: Robust Deep Combined Computing Tackling Heavy-Tailed Distribution in Medical Images},
  author={Liu, Yuanye and Zhen, Ruoxuan and Gao, Shangqi and Luo, Xinzhe and Gao, Xin and Chen, Qingchao and Zhuang, Xiahai},
  booktitle=MICCAI,
  pages={45--54},
  year={2025},
}

@article{J_2021_domain_shift,
  title={Domain adaptation for medical image analysis: a survey},
  author={Guan, Hao and Liu, Mingxia},
  journal={IEEE Transactions on Biomedical Engineering},
  volume={69},
  number={3},
  pages={1173--1185},
  year={2021},
  publisher={IEEE}
}

@inproceedings{C_2025ICML_forgetting,
  title={Predicting the Susceptibility of Examples to Catastrophic Forgetting},
  author={Hacohen, Guy and Tuytelaars, Tinne},
  booktitle=ICML,
  year = {2025}
}

@inproceedings{C_2024NIPS_LWE,
  title={Learning where to edit vision Transformers},
  author={Yang, Yunqiao and Huang, Long-Kai and Chen, Shengzhuang and Ma, Kede and Wei, Ying},
  booktitle=NIPS,
  volume={37},
  pages={134459--134487},
  year={2024}
}

@inproceedings{C_2025NIPS_ViT_edit,
  title={Model Editing for Vision Transformers},
  author={Huang, Xinyi and Zhao, Kangfei and Huang, Long-Kai},
  booktitle=NIPS,
  year={2025}
}

@inproceedings{C_2023ICLR_MEMIT,
 title={Mass-editing memory in a transformer},
  author={Meng, Kevin and Sharma, Arnab Sen and Andonian, Alex and Belinkov, Yonatan and Bau, David},
    booktitle = ICLR,
    year = 2023
}

@inproceedings{C_2024ACLf_wilke,
  title={Wilke: Wise-layer knowledge editor for lifelong knowledge editing},
  author={Hu, Chenhui and Cao, Pengfei and Chen, Yubo and Liu, Kang and Zhao, Jun},
  booktitle={Findings of the Association for Computational Linguistics (ACL)},
  pages={3476--3503},
  year={2024}
}

@article{J_2017NAC_EWC,
  title={Overcoming catastrophic forgetting in neural networks},
  author={Kirkpatrick, James and Pascanu, Razvan and Rabinowitz, Neil and Veness, Joel and Desjardins, Guillaume and Rusu, Andrei A and Milan, Kieran and Quan, John and Ramalho, Tiago and Grabska-Barwinska, Agnieszka and others},
  journal={Proceedings of the national academy of sciences},
  volume={114},
  number={13},
  pages={3521--3526},
  year={2017},
}

@article{medmnistv2,
    title={MedMNIST v2-A large-scale lightweight benchmark for 2D and 3D biomedical image classification},
    author={Yang, Jiancheng and Shi, Rui and Wei, Donglai and Liu, Zequan and Zhao, Lin and Ke, Bilian and Pfister, Hanspeter and Ni, Bingbing},
    journal={Scientific Data},
    volume={10},
    number={1},
    pages={41},
    year={2023},
}

@article{J_2025MIA_merit,
  title = {MERIT: Multi-view evidential learning for reliable and interpretable liver fibrosis staging},
  author={Liu, Yuanye and Gao, Zheyao and Shi, Nannan and Wu, Fuping and Shi, Yuxin and Chen, Qingchao and Zhuang, Xiahai},
  journal = {Medical Image Analysis},
  volume = {102},
  pages = {103507},
  year = {2025},
}

@article{J_2019NMI_continual_owm,
  title={Continual learning of context-dependent processing in neural networks},
  author={Zeng, Guanxiong and Chen, Yang and Cui, Bo and Yu, Shan},
  journal={Nature Machine Intelligence},
  volume={1},
  number={8},
  pages={364--372},
  year={2019},
}

@article{J_2025BMJ_future-ai,
  title={FUTURE-AI: International consensus guideline for trustworthy and deployable artificial intelligence in healthcare},
  author={Lekadir, Karim and Frangi, Alejandro F and Porras, Antonio R and Glocker, Ben and Cintas, Celia and Langlotz, Curtis P and Weicken, Eva and Asselbergs, Folkert W and Prior, Fred and Collins, Gary S and others},
  journal={bmj},
  volume={388},
  year={2025},
  publisher={British Medical Journal Publishing Group}
}

@article{J_2022MIA_inter_survey,
  title={Explainable artificial intelligence (XAI) in deep learning-based medical image analysis},
  author={Van der Velden, Bas HM and Kuijf, Hugo J and Gilhuijs, Kenneth GA and Viergever, Max A},
  journal={Medical image analysis},
  volume={79},
  pages={102470},
  year={2022},
}

@article{J_2023NBE_fairness,
  title={Algorithmic fairness in artificial intelligence for medicine and healthcare},
  author={Chen, Richard J and Wang, Judy J and Williamson, Drew FK and Chen, Tiffany Y and Lipkova, Jana and Lu, Ming Y and Sahai, Sharifa and Mahmood, Faisal},
  journal={Nature biomedical engineering},
  volume={7},
  number={6},
  pages={719--742},
  year={2023},
}

@string{ICML = "Proceedings of the International Conference on Machine Learning (ICML)"}

@string{NIPS = "Proceedings of the Annual Conference on Neural Information Processing Systems (NeurIPS)"}

@string{ICLR = "Proceedings of the International Conference on Learning Representations (ICLR)"}

@string{MICCAI = "Proceedings of the International Conference on Medical Image Computing and Computer Assisted Intervention (MICCAI)"}
\end{document}